\documentclass[pdflatex,sn-nature]{sn-jnl}
\usepackage{geometry}
\usepackage{array}

\usepackage{graphicx}%
\usepackage{multirow}%
\usepackage{amsmath,amssymb,amsfonts}%
\usepackage{amsthm}%
\usepackage{mathrsfs}%
\usepackage[title]{appendix}%
\usepackage{xcolor}%
\usepackage{textcomp}%
\usepackage{manyfoot}%
\usepackage{booktabs}%
\usepackage{algorithm}%
\usepackage{algorithmicx}%
\usepackage{algpseudocode}%
\usepackage{listings}%
\usepackage{float}
\usepackage{subcaption}
\usepackage{tabularx}

\theoremstyle{thmstyleone}%

\theoremstyle{thmstyletwo}%

\theoremstyle{thmstylethree}%

\unnumbered

\begin{document}

\title[Autonomous discovery of traffic laws with AI traffic scientists]{Autonomous discovery of traffic laws with AI traffic scientists}

\author[1]{\fnm{Xingyuan} \sur{Dai}}
\equalcont{These authors contributed equally to this work.}

\author[1,2]{\fnm{Yue} \sur{Liu}}
\equalcont{These authors contributed equally to this work.}

\author[1]{\fnm{Xiaoyan} \sur{Gong}}
\equalcont{These authors contributed equally to this work.}

\author[2]{\fnm{Qinghai} \sur{Miao}}
\equalcont{These authors contributed equally to this work.}

\author[1,2]{\fnm{Junyou} \sur{Shang}}

\author[1]{\fnm{Yutong} \sur{Wang}}

\author[1]{\fnm{Chao} \sur{Guo}}

\author[1]{\fnm{Yonglin} \sur{Tian}}

\author[1,2]{\fnm{Yizhang} \sur{Chai}}

\author[3]{\fnm{Chao} \sur{Xiang}}

\author*[1,4]{\fnm{Yisheng} \sur{Lv}}\email{yisheng.lv@ia.ac.cn}
\author*[4,5]{\fnm{Fei-Yue} \sur{Wang}}\email{feiyue.wang@ia.ac.cn}

\affil[1]{State Key Laboratory of Multimodal Artificial Intelligence Systems, Institute of Automation, Chinese Academy of Sciences, Beijing 100190, China}
\affil[2]{School of Artificial Intelligence, University of Chinese Academy of Sciences, Beijing 100049, China}
\affil[3]{China Telecom Research Institute, Beijing, 102209, China}
\affil[4]{Macau Institute of Systems Engineering, Macau University of Science and Technology, Macao 999078, China}
\affil[5]{State Key Laboratory for Management and Control of Complex Systems, Institute of Automation, Chinese Academy of Sciences, Beijing 100190, China}

\abstract{
Universal traffic laws describe recurrent patterns in congestion, mobility and driving behavior across cities, providing a scientific basis for transportation planning, management and control. Their discovery, however, remains expert-driven, requiring candidate regularities to be identified from heterogeneous observational evidence or validated through intervention experiments. Although autonomous artificial intelligence (AI) systems have advanced scientific discovery in controlled laboratory settings, extending them to complex transportation domains remains a challenge. Here we present TrafficSci, an agentic AI system that formulates traffic-law discovery as an iterative, auditable workflow integrating evidence scoping, critic--judge hypothesis induction, and observational--interventional validation. Across four case studies spanning population, network, control and trajectory scales, TrafficSci autonomously rediscovers three established traffic laws and identifies an unreported intrinsic temporal memory scale in urban driving behavior, statistically consistent across eight cities and two trajectory datasets. TrafficSci provides a route for extending AI-driven scientific discovery from controlled domains to complex urban systems.
}

\maketitle

Urban transportation networks are among the most complex systems that cities should actively manage. Decisions on congestion mitigation, signal deployment, route guidance and long-term infrastructure investment all rest, ultimately, on quantitative understanding of how traffic behaves: how congestion costs distribute across a network, how populations allocate trips over space and time, how individual drivers respond to preceding conditions, and how the benefits of a control technology scale with its deployment level. Such understanding is encoded in traffic laws, concise and testable regularities that recur across operating conditions and, crucially, across cities \cite{duan2023spatiotemporal,saberi2020simple,case2,cabanas-tirapu2025human,vazifeh2018addressing}. These laws provide interpretable priors for prediction, principled boundaries for control design, and transferable foundations for urban planning \cite{hamedmoghadam2021percolation,trafficevacuation,olmos2018collapse,paralleltraffic}.  
However, the set of established traffic laws remains limited relative to the complexity of urban transportation systems.
Data-driven prediction and control models have advanced at a remarkable pace \cite{traiffcflow}, but the interpretable scientific regularities on which robust traffic management ultimately depends have not kept step, leaving a widening gap between the field's capacity to observe and optimize increasingly complex traffic systems and its capacity to explain why particular interventions work and where their limits lie.

The difficulty of closing this gap is both practical and structural. A conventional traffic-law study begins with manual literature synthesis, proceeds through iterative variable and metric design, and demands substantial implementation effort, including data selection, cleaning and bespoke analysis code, before a single hypothesis can be tested \cite{symbolic_sci}. Credible validation typically requires checks across multiple datasets, time periods and operating regimes; questions involving management interventions further require controlled simulation experiments. Results frequently prompt revisions that restart the entire cycle. These practical burdens are compounded by a fundamental asymmetry with laboratory sciences. Urban traffic systems are open, non-stationary and tightly coupled to human behavior; no controlled physical experiment can be arranged to isolate variables and probe candidate laws directly. Every hypothesis should therefore be corroborated through observational data analysis, computational simulation, or a combination of both. This dual evidentiary requirement, together with the high manual cost of each experimental iteration, confines the hypothesis space that any individual research group can explore within a realistic time frame and leaves potentially important traffic regularities undiscovered.

Recent advances in large language models (LLMs) have opened a route to accelerating scientific discovery by organizing research as executable, verifiable workflows \cite{wang2023scientific,natcities_llm_planning,zhang2025exploring,chen2024navigating, airesearch}. In chemistry, the Coscientist system integrates language models with robotic tools to plan and execute experiments autonomously \cite{boiko2023autonomous}; in materials science, A-Lab couples computational prediction with robotic synthesis to discover and produce novel compounds within days \cite{szymanski2023autonomous};
in biomedical science, 
Co-Scientist further demonstrates the potential of LLM agents in scientific discovery by integrating literature reasoning, hypothesis generation, experimental planning, and iterative refinement into an automated research workflow \cite{CoScientist}. These successes share a structural prerequisite: a closed loop in which hypothesis generation is tightly coupled to experimental validation \cite{llm4trans}. In each case, the experimental side of the loop relies on controllable laboratory apparatus. Whether a comparable loop can be constructed for domains where controlled experiments are often unavailable, and where candidate laws should survive both observational and interventional scrutiny, has not been systematically explored in traffic science.

In this paper, we introduce TrafficSci, an agentic AI system designed to close this discovery loop for urban transportation science. Given a research topic, TrafficSci retrieves and organizes domain literature through a structured tree-search mechanism, constructs candidate traffic laws as falsifiable hypotheses anchored to cited evidence, and validates them through an experimentation module that supports both real-world data analysis and simulation-based intervention testing \cite{mathematical}. When test outcomes fail to corroborate a candidate, the system revises the hypothesis and re-enters the validation cycle. The central design principle is that observational corroboration and interventional testing operate within a single iterative loop, addressing the dual evidentiary standard that a field without controllable laboratories inherently requires.

We evaluate TrafficSci through four case studies spanning population-level mobility and visitation scaling, network-level congestion dynamics, control-oriented intervention evaluation, and trajectory-level temporal regularity in driving.  Across the first three case studies, TrafficSci autonomously rediscovers and empirically verifies the established traffic laws reported in prior work, without manual specification of hypotheses or validation procedures.
Beyond rediscovery, TrafficSci discovers a previously unreported law, a stable temporal memory-scale regularity in driving behavior that is identified directly from large-scale vehicle trajectories without predefined hypotheses or analytical templates and remains consistent across cities. 
Taken together, these results indicate that traffic-law discovery can be organized as a repeatable, auditable computational process that recovers established knowledge and identifies regularities missed by conventional research, 
offering an initial demonstration that AI-driven scientific discovery can extend from controlled laboratories to complex urban transportation systems.
\section{Results}\label{sec2}

\subsection{System overview}
\begin{figure}[t]
    \centering
    \includegraphics[width=0.98\textwidth]{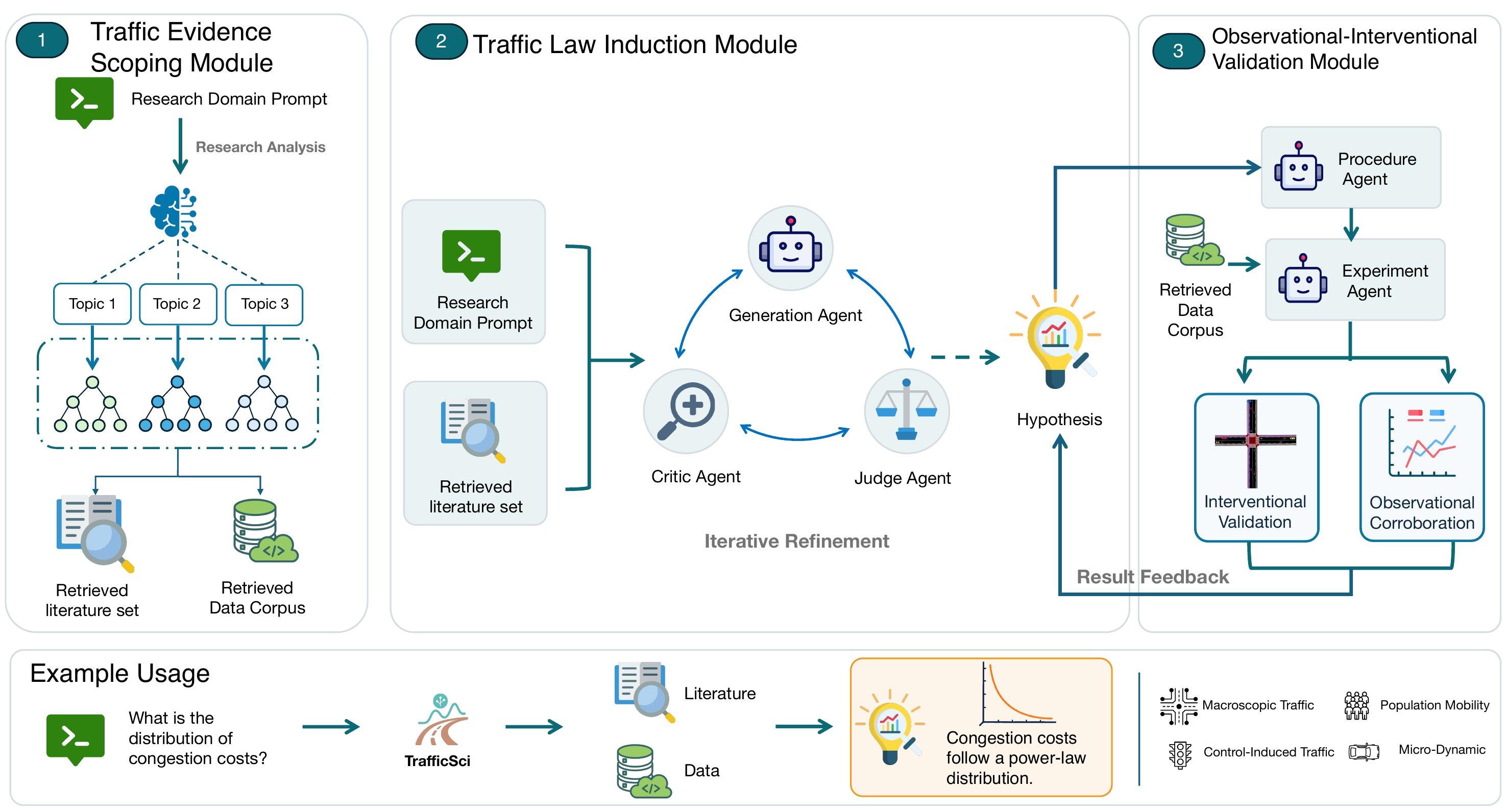}
    \caption{The architecture of TrafficSci. TrafficSci begins with the traffic evidence scoping module, which retrieves relevant literature and constructs an evidence corpus. The traffic law induction module then formulates structured hypotheses from the retrieved evidence. These hypotheses are evaluated by the observational--interventional validation module through statistical observation on real-world traffic data or intervention-based experiments in simulation environments. The resulting experimental evidence is fed back to update and refine the hypotheses, forming a closed-loop process for automated discovery in transportation science.}
    \label{fig:framework}
\end{figure}

As shown in Fig.~\ref{fig:framework}, TrafficSci is an autonomous system designed for the discovery and verification of traffic laws, specifically tailored to the unique characteristics of urban transportation research. 
It follows an agentic AI-driven workflow comprising paper retrieval, hypothesis construction, automated experimentation, and feedback-driven hypothesis refinement within the induction module. 
The collaboration between these agents enables TrafficSci to discover and validate traffic laws across population-level mobility, network-level congestion, control-oriented intervention, and trajectory-level driving behavior.
To make this workflow operational, TrafficSci organizes the discovery process into three interacting functional modules:
\begin{itemize}
    \item \textbf{The Traffic Evidence Scoping Module} uses the Literature-based Agent Tree Search (Lit-LATS) framework to autonomously retrieve relevant literature based on predefined transportation topics. It organizes the literature and extracts key information, forming a structured knowledge base that informs hypothesis construction and the discovery of traffic laws.
    \item \textbf{The Traffic Law Induction Module} generates structured, testable hypotheses based on the knowledge base from the traffic evidence scoping module. It defines traffic variables, their relationships, and the conditions under which they apply, providing hypotheses for experimental validation.

    \item \textbf{The Observational--Interventional Validation Module} designs and executes experiments for hypotheses generated by the traffic law induction module. It performs observational validation on real-world traffic data or interventional validation in simulation environments where traffic conditions can be actively manipulated. Supported by an extensible transportation database, the module feeds results back to refine, reject, or generalize hypotheses in a closed-loop framework.
\end{itemize}

\begin{figure*}[!t]
    \centering
    \includegraphics[width=\textwidth]{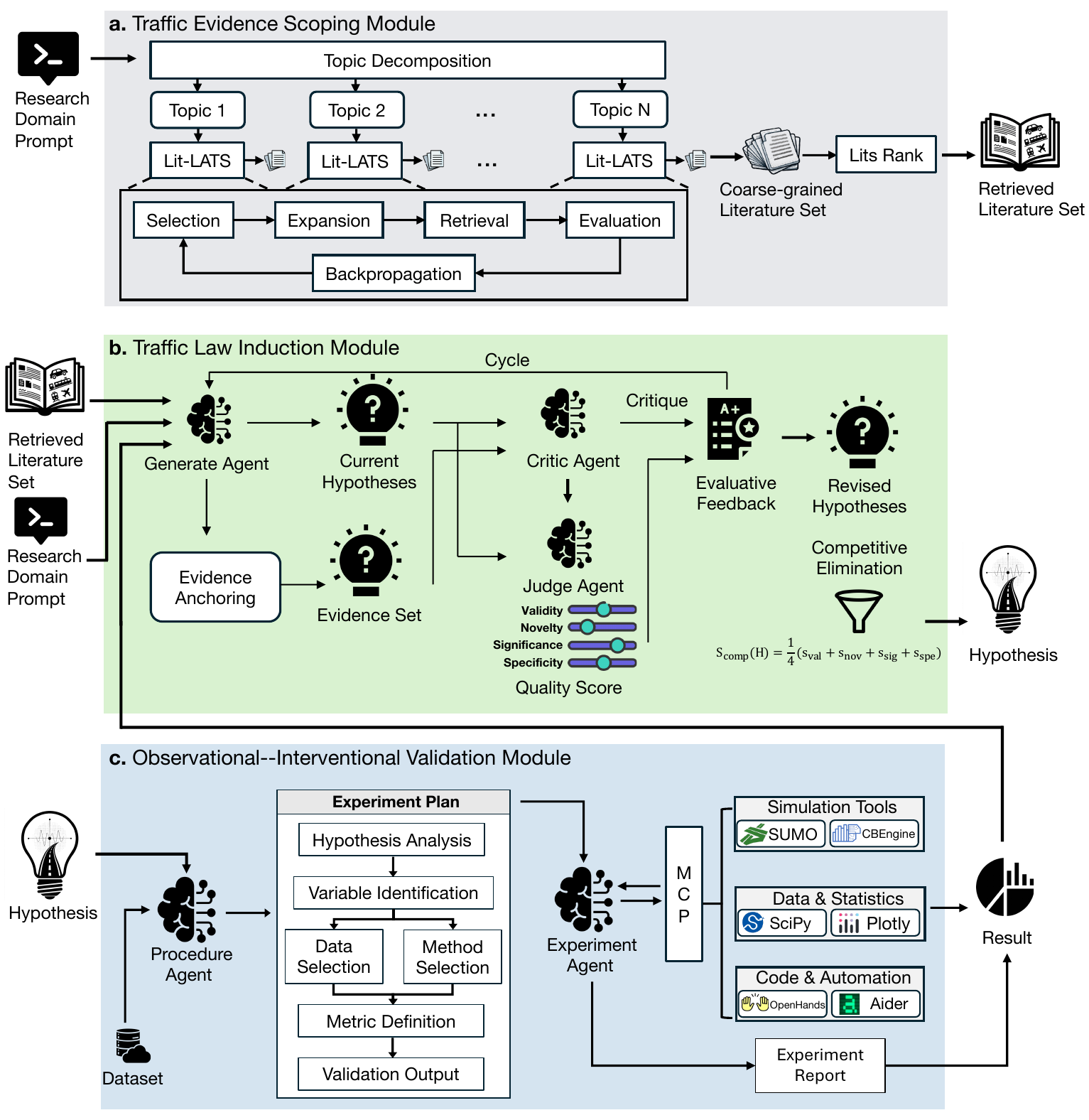}
    \caption{
    \textbf{Detailed workflows of TrafficSci.}
    \textbf{a} \textbf{Traffic evidence scoping module.}
    Given a research domain prompt, the system performs topic decomposition and applies Lit-LATS to each sub-topic for structured literature exploration, including selection, expansion, retrieval, evaluation, and backpropagation, producing a coarse-grained and ranked literature set.
    \textbf{b} \textbf{Traffic law induction module.}
    Using the retrieved literature and domain prompt, a generation agent proposes structured candidate hypotheses with explicit evidence anchoring.
    A critic--judge loop screens candidate hypotheses in terms of validity, conceptual novelty, significance, and specificity before competitive elimination.
    The selected hypothesis is then passed to the validation module together with its evidence set, critique record, and validation-route tag.
    \textbf{c} \textbf{Observational--interventional validation module.}
    Refined hypotheses are translated into executable procedures by a procedure agent and empirically tested by an experiment agent via MCP-based tool interaction (e.g., SUMO, SciPy, and OpenHands), producing structured experimental results.
    }
    \label{fig:TrafficSci_detailed_design}
\end{figure*}

Together, these modules transform a high-level research topic into empirically tested traffic-law hypotheses, with their detailed workflows illustrated in Fig.~\ref{fig:TrafficSci_detailed_design}.

\subsection{Benchmarking rediscovery of established traffic laws}

We use three established urban traffic laws as benchmark rediscovery tasks to evaluate whether TrafficSci can reconstruct testable hypotheses and validation procedures from earlier evidence. 
Three representative cases are selected to cover different scales of traffic phenomena, including human mobility, congestion dynamics, and control-induced traffic dynamics. To ensure that the rediscovery process does not rely on direct access to the target studies, the literature retrieval stage is restricted to papers published before the corresponding reference paper.

\begin{figure}[!htbp]
    \centering
    \includegraphics[width=1.0\textwidth]{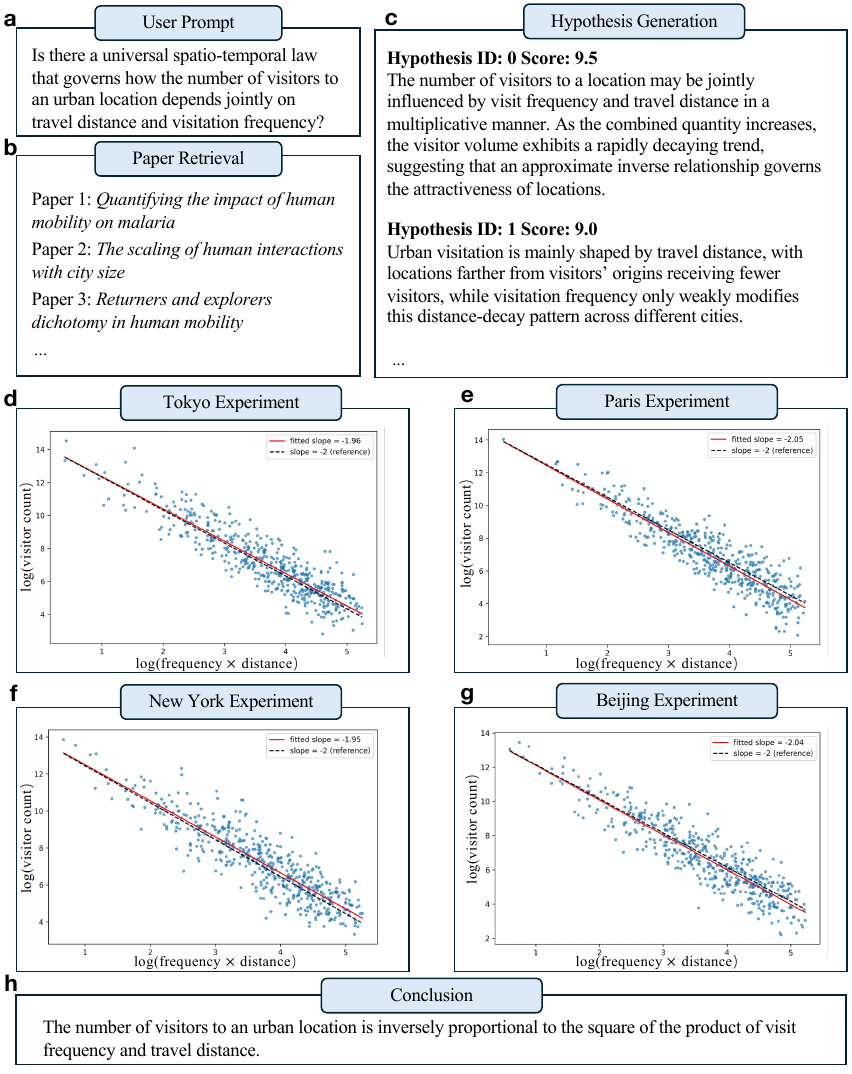}
    \caption{\textbf{Visualization of how TrafficSci discovers the spatio-temporal law governing visitor volume at urban locations as a joint function of travel distance and visitation frequency.}
    \textbf{a} User-provided prompt specifying the target scientific question.
    \textbf{b} Literature retrieval process for identifying relevant prior studies and empirical evidence.
    \textbf{c} Construction of testable hypotheses from the retrieved literature.
    \textbf{d--g}  Log--log relationships between visitor volume and the combined distance-frequency variable in Tokyo, Paris, New York and Beijing.    \textbf{h} System-generated conclusion. The results demonstrate a consistent inverse-square scaling relationship across cities.}
    \label{fig:exper2}
\end{figure}

\subsubsection{Universal visitation law of human mobility across cities}

Human mobility in cities is characterized by repeated visits to diverse locations, forming rich spatio-temporal visitation patterns across urban environments \cite{urban_ride}. A central question is whether visitor volume can be explained by a general law that jointly accounts for travel distance and visitation frequency, rather than by spatial distance alone. This issue is closely related to the understanding of recurrent population flows, urban interaction intensity, and location demand. The study \textit{The Universal Visitation Law of Human Mobility} \cite{case2} reported that the number of visitors to a given location scales inversely with the square of the product of travel distance and visitation frequency, revealing a compact spatio-temporal relation that remains stable across heterogeneous cities \cite{limits,natcities_mobility_expansion}.

To revisit this law, TrafficSci organizes the problem around three core variables---travel distance, visitation frequency, and visitor volume---and links literature-grounded hypothesis generation with multi-city empirical validation. As illustrated in Fig.~\ref{fig:exper2}, the system retrieves relevant studies, proposes candidate functional forms, and then tests them on mobility data from Tokyo, Paris, New York and Beijing. The resulting log--log plots recover the same inverse-square scaling pattern across all four cities, in agreement with the published universal visitation law.

\subsubsection{Congestion cost distribution in urban and suburban areas based on jam-prints}

\begin{figure}[!htbp]
    \centering
    \includegraphics[width=1.0\textwidth]{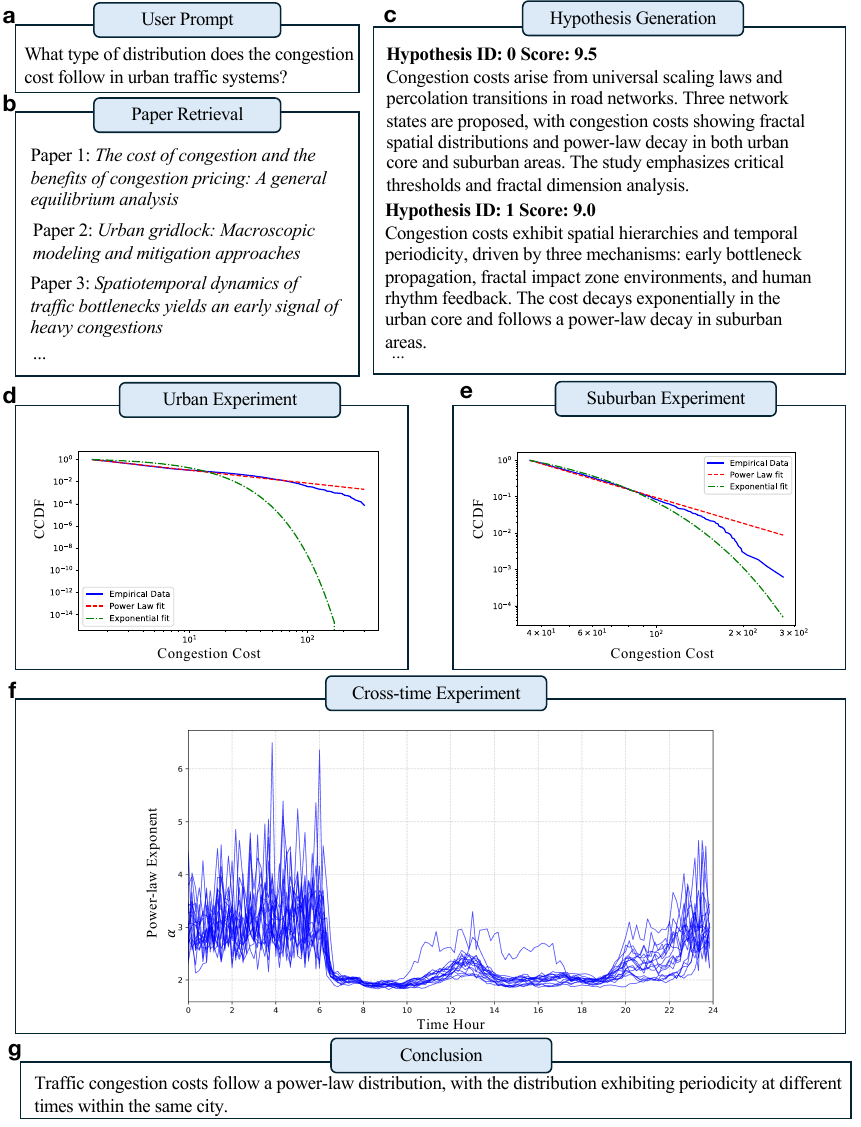}
    \caption{\textbf{Visualization of the TrafficSci process for discovering the distribution of traffic congestion costs.}
    \textbf{a} User-provided prompt specifying the target scientific question.
    \textbf{b} Literature retrieval process for identifying relevant prior studies and empirical evidence.
    \textbf{c} Construction of testable hypotheses from the retrieved literature.
    \textbf{d} Experimental validation of congestion cost distribution in urban areas.
    \textbf{e} Experimental validation of congestion cost distribution in suburban areas.
    \textbf{f} Verification of intra-city periodicity of congestion cost distribution across multiple days.
    The results confirm that traffic congestion costs follow a power-law distribution and exhibit stable temporal regularity within the same city.}
    \label{fig:exper1}
\end{figure}

Urban congestion strongly affects traffic efficiency and quality of life, and its impact can be characterized through the distribution of congestion costs, such as delay and fuel consumption \cite{ccolak2016understanding}. A natural scientific question is whether these costs follow a reproducible statistical law rather than varying in an arbitrary manner across time and space. In \textit{Unveiling City Jam-Prints of Urban Traffic Based on Jam Patterns} \cite{case1}, it was found that congestion cost distributions in both urban and suburban areas exhibit a power-law form, implying that a small number of jam events account for disproportionately large costs \cite{jamming_transition,cogshapecity,natcities_congestion_routing,movepatterns}. The same study further showed that the corresponding scaling exponents remain similar across days and across the same hours on different days within a city, forming distinctive jam-print patterns.

TrafficSci revisits this problem by combining literature-guided hypothesis construction with empirical analysis on urban and suburban traffic data. As presented in Fig.~\ref{fig:exper1}, the system retrieves prior studies on congestion cost patterns, proposes candidate hypotheses about distributional form and temporal variation, and validates them through targeted experiments. The recovered results confirm the power-law distribution of congestion costs and further reproduce the periodic intra-city consistency of the scaling behavior, indicating that recurrent congestion patterns form a stable spatial signature of urban traffic.
\begin{figure}[!htbp]
    \centering
    \includegraphics[width=1.0\textwidth]{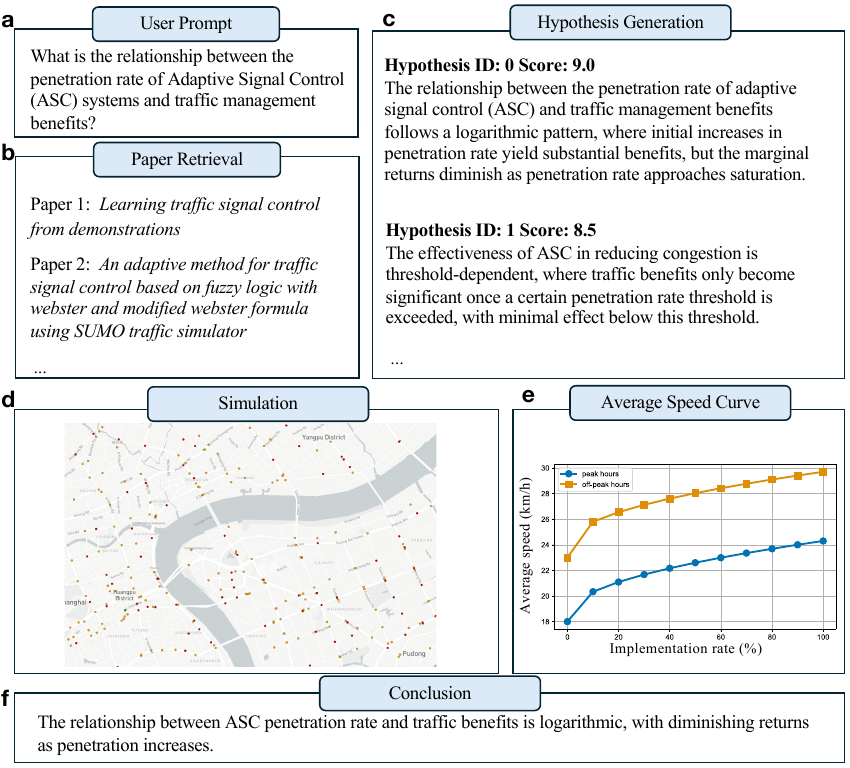}
    \caption{\textbf{Autonomous discovery of the logarithmic relationship between ASC penetration and traffic benefits.}
    \textbf{a} User-provided prompt specifying the target scientific question.
    \textbf{b} Literature retrieval process for identifying relevant prior studies and empirical evidence.
    \textbf{c} Construction of testable hypotheses from the retrieved literature.
    \textbf{d} Traffic simulation illustrating the experimental setup and system-level effects under different ASC penetration rates.
    \textbf{e} Relationship between ASC penetration rate and traffic benefits under peak and off-peak conditions, both exhibiting a logarithmic scaling pattern.}
    \label{fig:case4_composite}
\end{figure}

\subsubsection{Logarithmic relationship between adaptive signal control penetration rate and traffic management benefits}

Adaptive signal control (ASC) is widely recognized as an effective strategy for reducing congestion and improving traffic efficiency \cite{asc_nature,asc_timc}. Yet an important deployment question remains: how do traffic management benefits change as ASC penetrates an urban network? This question matters not only for understanding the scaling behavior of intelligent traffic systems, but also for informing cost-effective deployment decisions \cite{flow}. In \textit{Big-data Empowered Traffic Signal Control Could Reduce Urban Carbon Emission} \cite{case4}, the reported results showed that the penetration--benefit relationship follows a logarithmic trend, with rapid gains at low penetration levels and progressively weaker marginal returns as deployment expands.

TrafficSci examines this problem through simulation-based interventional validation, where the ASC penetration rate is actively controlled as the intervention variable. The system reproduces the experiments using the same CBEngine simulation platform and a consistent evaluation protocol. As shown in Fig.~\ref{fig:case4_composite}, TrafficSci retrieves relevant studies, formulates a diminishing-return hypothesis, and then conducts controlled simulation experiments by systematically varying the ASC penetration rate. The resulting changes in travel time, congestion level, traffic efficiency, and carbon-related indicators are recorded under both peak and off-peak conditions. These intervention-response curves consistently reproduce the logarithmic penetration--benefit pattern reported in the reference study.

Across the three benchmarks, TrafficSci recovered each published law without manually specifying the hypotheses or procedures: the inverse-square visitation scaling and the power-law congestion-cost distribution by observational validation, and the logarithmic relationship between ASC penetration and traffic benefits by controlled intervention.

\subsection{Discovery of an intrinsic temporal memory scale in urban driving behavior}

\subsubsection{Motivation and autonomous hypothesis generation}

A fundamental question in traffic science is whether microscopic driving behavior contains stable temporal regularities that can be measured and generalized across urban environments \cite{saifuzzaman2014,huang2018,ma2020}. Existing trajectory prediction and traffic simulation studies widely use historical motion states as model inputs, implicitly assuming that recent driving history provides useful information for future behavior. However, the temporal dependence itself is usually treated as a predefined modeling choice, such as a fixed observation window or a manually selected history length, rather than as an object of scientific discovery. As a result, it remains unclear whether individual driving behavior possesses an intrinsic temporal memory scale, how long the influence of past states persists, and whether this temporal structure is consistent across different cities.

This question is important because the temporal memory of driving behavior reflects how drivers respond to recent motion states, surrounding constraints, and evolving traffic interactions. If such memory has a stable statistical form, it may provide a microscopic behavioral law that complements macroscopic traffic regularities such as flow-density relationships and mobility scaling laws.

To investigate this problem in an open-ended manner, TrafficSci is provided with a high-level scientific inquiry: whether individual driving behavior exhibits statistically stable temporal dependence across heterogeneous urban environments. No candidate metric, predefined formula, or expected distribution is specified in advance. Starting from this inquiry, the traffic evidence scoping module retrieves prior concepts related to temporal dependence, memory effects, correlation decay, behavioral persistence, and human mobility regularities. These retrieved studies provide conceptual evidence that historical states may influence current behavior, but they do not directly prescribe a specific measurable law for urban driving trajectories.

Based on the retrieved evidence, the traffic law induction module further formulates a testable hypothesis: if urban driving behavior contains an intrinsic temporal structure, then the influence of historical driving states on current states should decay as the temporal lag increases, and the effective decay horizon can be quantified as a temporal memory scale. This hypothesis transforms the original abstract question into an empirically testable proposition. In this process, the proposed quantity $\tau$ is not manually imposed by researchers, but emerges from the autonomous workflow of literature-grounded reasoning, hypothesis generation, and experimental design.

\begin{figure}[!t]
    \centering
    \includegraphics[width=1.0\textwidth]{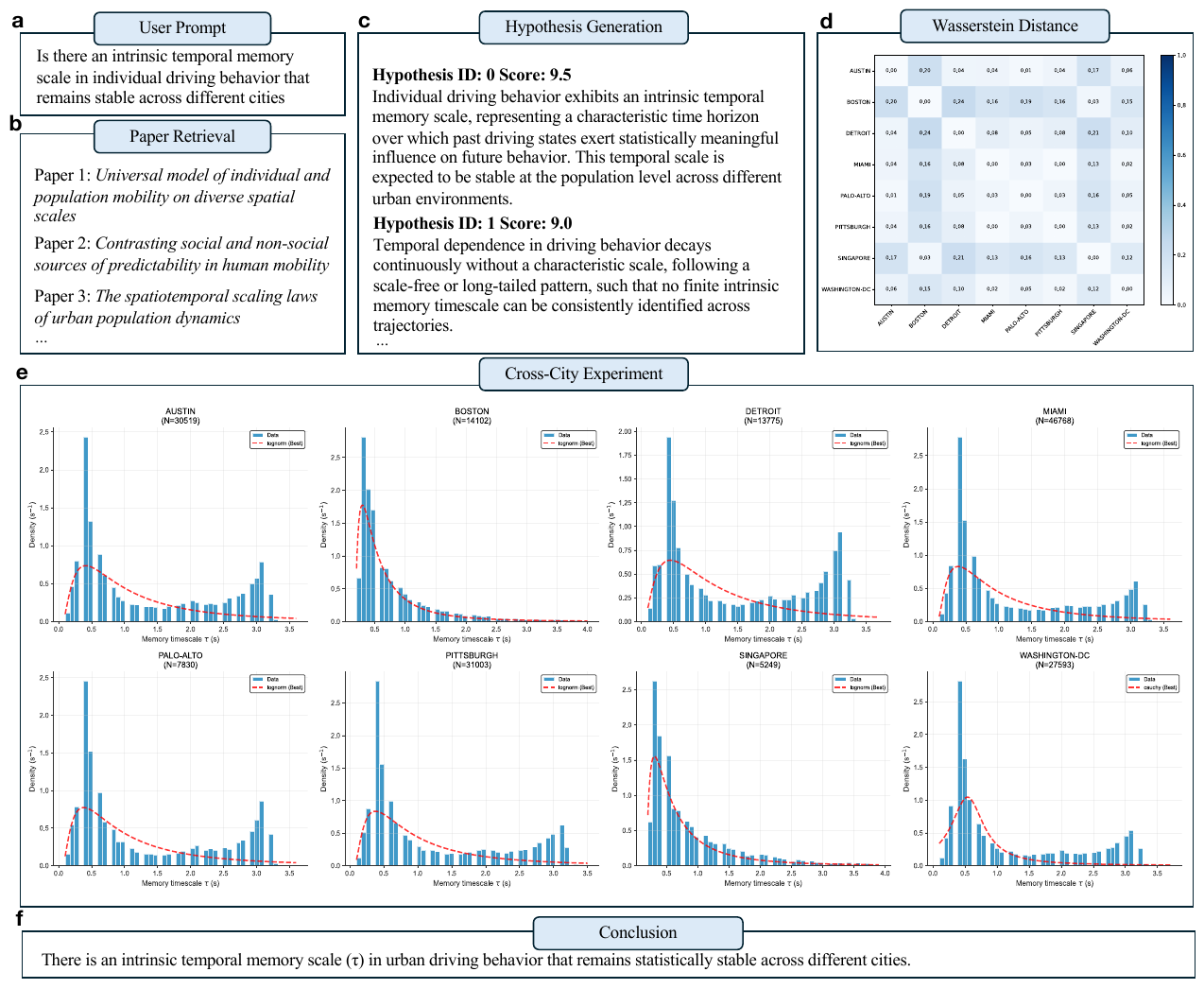}
    \caption{\textbf{Autonomous discovery of the intrinsic temporal memory scale in urban driving behavior.}
    \textbf{a} User-provided prompt specifying the target scientific question.
    \textbf{b} Literature retrieval process for identifying relevant prior studies and empirical evidence.
    \textbf{c} Construction of testable hypotheses from the retrieved literature.
    \textbf{d} Pairwise Wasserstein distances between city-level $\tau$ distributions. Darker colors indicate larger distributional distances.
    \textbf{e} City-level distributions of $\tau$, showing similar temporal-memory patterns across heterogeneous urban environments.
    \textbf{f} System-generated conclusion. The consistently small Wasserstein distances support the cross-city stability of the discovered memory scale.}
    \label{fig:case3_composite}
\end{figure}

\subsubsection{Definition and estimation of temporal memory scale}

To test the above hypothesis, the abstract concept of temporal dependence is first converted into a measurable quantity. TrafficSci therefore operationalizes the temporal memory of driving behavior from microscopic vehicle trajectories. For each vehicle at time $t$, a driving-state vector is constructed as
\begin{equation}
\mathbf{s}_t = \left[x_t,\; y_t,\; v_t,\; \theta_t\right]^\top,
\end{equation}
where $(x_t,y_t)$ denotes the vehicle position in the ground-plane coordinate system, $v_t$ denotes the instantaneous speed, and $\theta_t$ denotes the heading angle. This representation jointly describes spatial displacement, motion intensity, and directional evolution, which are the basic elements of short-term driving behavior. Since vehicle positions are naturally continuous over time, the position terms are used to describe the observable kinematic continuity of microscopic trajectories, while the speed and heading terms provide complementary information about motion intensity and directional evolution. Accordingly, the estimated $\tau$ is interpreted as a temporal dependence scale of driving-state evolution, rather than as a direct measure of cognitive driver memory.

Given the state sequence, TrafficSci estimates how strongly historical states remain statistically related to the current driving state over time.
Specifically, for a temporal lag $\Delta$, the lag-dependent historical influence function is defined as
\begin{equation}
I(\Delta)=\frac{1}{4}\left(
\rho_x(\Delta)+\rho_y(\Delta)+\rho_v(\Delta)+\rho_\theta(\Delta)
\right),
\end{equation}
where $\rho_x(\Delta)$, $\rho_y(\Delta)$, $\rho_v(\Delta)$, and $\rho_\theta(\Delta)$ denote the lag-$\Delta$ autocorrelation coefficients of the position ($x$, $y$), speed ($v$), and heading ($\theta$) components. Intuitively, $I(\Delta)$ measures the remaining statistical influence of historical driving states when the time interval from the current state increases. A larger value of $I(\Delta)$ indicates stronger dependence on the past, whereas a value close to zero suggests that the influence of the corresponding historical state has largely decayed.

Based on this influence function, the temporal memory scale is defined as the earliest lag at which the historical influence becomes sufficiently weak:
\begin{equation}
\tau=\min \left\{ \Delta > 0 \;\middle|\; |I(\Delta)| < \epsilon \right\},
\end{equation}
where $\epsilon=0.05$ is used as a weak-correlation threshold to determine when the temporal dependence becomes negligible. This choice is consistent with the interpretation that correlation magnitudes close to zero indicate negligible association \cite{correlation}, suggesting that the historical influence has largely decayed. This threshold serves as an operational criterion for identifying the decay point, and the same value is consistently applied across all cities and datasets to ensure comparable estimates of $\tau$. Under this definition, $\tau$ represents the effective temporal horizon over which past driving states still exert observable influence on current behavior.

It is worth emphasizing that $\tau$ is not a manually selected observation length or a hyperparameter of a prediction model. Instead, it is estimated directly from the empirical temporal correlation structure of real-world trajectories.

\subsubsection{Cross-city validation of the discovered law}

After estimating the temporal memory scale $\tau$, TrafficSci further examines whether the discovered quantity
reflects a stable behavioral regularity rather than a city-specific artifact. To this end, we conduct cross-city
validation on large-scale vehicle trajectories from six cities in Argoverse 2 \cite{Argoverse2} and two cities in nuScenes \cite{nuScenes}. 
Spanning multiple U.S. cities and Singapore, the two datasets differ in collection protocol, temporal resolution, scenario duration, and sensor configuration, and together form a complementary, cross-source testbed for evaluating whether $\tau$ generalizes across heterogeneous urban environments rather than reflecting a single collection pipeline.

Fig.~\ref{fig:case3_composite}e shows the empirical distributions of $\tau$ across different cities. Although the trajectories are collected from distinct urban environments, the distributions exhibit a highly consistent pattern. Most samples are concentrated within a short-to-moderate temporal range, indicating that recent driving states dominate current behavioral decisions. Meanwhile, all cities show a visible long-tail structure, suggesting that a subset of driving behaviors retains longer temporal dependence. This common distributional shape implies that the temporal memory scale captures a shared microscopic regularity of urban driving behavior.

To quantitatively evaluate the similarity between cities, we compute the normalized first-order Wasserstein distance between pairwise empirical distributions of $\tau$. As shown in Fig.~\ref{fig:case3_composite}d, across eight cities and two trajectory datasets, the temporal memory scale ranges from $0.0$ to $4.0$ s, with pairwise normalized Wasserstein distances below $0.24$ and bootstrap $95\%$ confidence intervals remaining below $0.10$, indicating limited distributional discrepancy among different urban environments. This result suggests that the discovered memory scale is not only observable within individual cities, but also exhibits strong cross-city consistency.
\subsubsection{Scientific significance and potential applications}

The temporal memory scale $\tau$ reframes the historical context of urban driving behavior from a fixed, manually chosen observation window into a measurable and reproducible quantity, revealing that the dependence of current behavior on the past has a stable temporal horizon that holds across cities. This carries three practical implications for urban traffic research. As a diagnostic, the empirical $\tau$ distribution can test whether a traffic simulator reproduces realistic microscopic temporal dynamics even when it already matches macroscopic statistics such as speed or density. As a transferable behavioral prior, the cross-city stability of $\tau$ can support simulator calibration, domain adaptation, and transfer of traffic models across urban environments. As a design guideline, $\tau$ offers an empirical basis for choosing the history length, memory modules, or state-history features in trajectory prediction and learning-based driving policies, rather than relying on heuristic windows. More broadly, this case shows how TrafficSci can move from an open-ended scientific question to a data-derived candidate regularity that warrants further validation across broader urban traffic conditions.

\section{Discussion}
This work shows that elements of traffic-law discovery can be organized as a closed-loop, auditable workflow in urban transportation science.  
TrafficSci coordinates multiple LLM agents to perform literature retrieval, hypothesis construction, automated experimentation and hypothesis evolution. We evaluate TrafficSci through four case studies spanning population-level visitation scaling \cite{case2}, network-level congestion \cite{case1}, infrastructure-induced intervention effects \cite{case4} and trajectory-level driving memory. These studies reveal four traffic regularities across distinct scales, including one previously unreported law that captures an intrinsic temporal memory scale in driving behavior.
These findings provide initial evidence that elements of traffic-law discovery can be systematized and automated in a reproducible manner.

From a methodological perspective, TrafficSci complements prevailing data-driven traffic modeling pipelines that prioritize prediction accuracy or control performance. Rather than treating scientific discovery as a secondary outcome of model fitting, TrafficSci explicitly treats transportation laws as first-class research objects. In this sense, TrafficSci functions as an AI traffic scientist that assists human researchers in formulating, testing, and refining traffic laws, thereby providing actionable scientific support for virtual-real parallel traffic management and control \cite{paralleltraffic}. Hypotheses are formulated in interpretable forms, linked to explicit empirical tests, and iteratively refined based on experimental feedback. This separation between law discovery and model optimization is particularly relevant for urban transportation science, where explanatory regularities often underpin city-level understanding, planning and policy analysis.

An important implication of this work is that traffic laws need not be restricted to closed-form mechanistic equations. In practice, traffic laws frequently manifest as statistical distributions, scaling relations or state-transition patterns that summarize collective system behavior across conditions. The evaluation cases illustrate this diversity by covering both network-level congestion phenomena under infrastructure and demand constraints, and mobility behaviors spanning individual and population scales. The ability of a single closed-loop agentic workflow to operate across these levels supports a unified view of transportation law discovery as a structured process that extracts candidate regularities, translates them into testable hypotheses, and validates or revises them through data-driven experiments.

Several limitations should be noted. TrafficSci depends on the availability and quality of existing literature and datasets, which may bias hypothesis generation or constrain empirical validation. In addition, the representativeness of available data limits the assessment of rare events or long-term structural changes. Although the closed-loop workflow reduces manual effort, scientific oversight remains necessary to ensure robustness when scaling automated discovery. Moreover, the cities and datasets examined here span a limited set of regions, so the global generalizability of the reported cross-city regularities, and their relevance across diverse urban contexts, remains to be established.

Future work may extend the framework to richer experimental environments and explore more interactive modes of human--AI collaboration. Overall, this study suggests that agentic scientific discovery offers a promising pathway toward more systematic, scalable and reproducible exploration of traffic laws, supporting the development of intelligent transportation systems and evidence-based urban planning.

\section{Methods}\label{sec_methods}
The overall framework of the TrafficSci system is illustrated in Fig.~\ref{fig:TrafficSci_detailed_design}. TrafficSci is designed as a closed-loop pipeline that enables automated discovery of urban transportation science hypotheses through literature retrieval, hypothesis construction, and experimental validation. TrafficSci is agnostic to the underlying language model and can be instantiated with any mainstream LLM, with all agents in this work implemented using GPT-5.5. The process begins with the traffic evidence scoping module, which systematically searches and filters traffic-related scientific literature across multiple topics. The retrieved studies are organized into a structured knowledge base, identifying key variables, empirical patterns, and commonly reported relationships, which provide direct knowledge support for hypothesis construction.
Based on this literature-grounded knowledge base, the traffic law induction module formulates explicit and testable traffic hypotheses by abstracting relationships among transportation variables and observed phenomena. These hypotheses are expressed in a structured form, enabling direct translation into experimental procedures.
The observational--interventional validation module validates the generated hypotheses by translating them into executable experimental workflows under two complementary paradigms. In observational validation, hypotheses are tested through data preprocessing and statistical analysis on real-world traffic datasets, such as vehicle trajectories, traffic flow records, congestion measurements, and road network information. In interventional validation, hypotheses are examined in traffic simulation environments such as SUMO, where key traffic conditions or control variables can be actively manipulated. The resulting experimental evidence is fed back to the hypothesis construction process, allowing hypotheses to be refined, rejected, or generalized based on empirical validation, thereby forming a closed-loop discovery pipeline.

\subsection{Traffic evidence scoping module}
In this work, the traffic evidence scoping module is primarily composed of a literature retrieval agent. The agent utilizes an enhanced Lit-LATS framework, which integrates a query generation mechanism driven by a language model with a topic exploration strategy based on Monte Carlo Tree Search (MCTS)  \cite{lats,aflow,tot}. Lit-LATS is specifically designed to efficiently acquire structured knowledge relevant to traffic research topics from the open literature space.

Traffic laws often manifest as statistical distributions, behavioral patterns, or cost variation relationships \cite{density,case1,case2}. As such, the associated literature tends to be scattered across various research areas, analysis scales, and application contexts. Moreover, the same traffic phenomenon may be described using different terminology and expressions in different papers. Traditional keyword-based retrieval methods are often inadequate in covering the diverse themes, scales, and heterogeneous expressions in such literature, leading to missed or fragmented clues about traffic laws. By incorporating a topic-exploration-based tree search mechanism, our traffic evidence scoping module dynamically expands the research topics related to traffic laws during the retrieval process. The system explores and selects among different topic branches, thereby ensuring more comprehensive coverage of the literature space required for traffic law discovery. This process provides a richer and more coherent knowledge base for subsequent hypothesis generation.

As shown in Fig.~\ref{fig:TrafficSci_detailed_design}a, the system starts by accepting a user-provided research topic, such as ``urban congestion cost distribution law'' or ``universal law of urban visitation patterns'', and passes it to the topic generation module. Using predefined LLM prompts, the system automatically generates multiple topic keywords. These keywords are denoted as
\begin{equation}
P = \{ p_1, p_2, \ldots, p_n \}.
\end{equation}

For each of these keywords, the system expands relevant domain-specific search terms to form a keyword set \( K_i = \{ k_{i1}, k_{i2}, \ldots, k_{im} \} \). The combination of each original keyword and its expanded search terms generates the following query set:
\begin{equation}
C = \{(p_i, k_{ij}) \mid p_i \in P,\ k_{ij} \in K_i\}.
\end{equation}

Each query combination is then used to retrieve related literature via the Semantic Scholar API, generating a candidate literature set. The abstracts of these documents are processed by the language model to assess whether the document contains quantitative descriptions of the research phenomenon, experimentally verifiable conclusions, or mechanistic explanations. Based on the relevance and heuristic value of the documents, a score \( Q \) is assigned to each keyword node, with a higher score indicating greater relevance to the research topic.

During the retrieval process, the system employs the Upper Confidence Bound for Trees (UCT) strategy from MCTS to determine whether a keyword node should continue expanding. For each node \( n \), the system maintains its cumulative score \( Q(n) \), which reflects the semantic relevance and heuristic value of the documents associated with that node. It also tracks the number of times the node has been visited \( N(n) \) and the number of visits to its parent node \( N(parent(n)) \). The node's expansion value is determined using the following UCT formula:

\begin{equation}
UCT(n) = Q(n) + c \cdot \sqrt{\frac{\ln N(parent(n))}{N(n)}},
\end{equation}
where \( Q(n) \) represents the cumulative score of the node, indicating its relevance to the research topic, \( N(n) \) is the number of visits to the node, \( N(parent(n)) \) is the number of visits to the parent node, and \( c \) is the exploration coefficient, which balances exploration and exploitation.

Through this process of topic keyword generation, keyword expansion, semantic retrieval, and the UCT-based node selection strategy, the system identifies a set of high-value literature nodes. These nodes satisfy the following criteria: they exhibit high semantic relevance to the research topic, contain quantitative descriptions, verifiable conclusions, or mechanistic explanations in the abstracts, and have high UCT scores during the expansion process.

To clarify the scope of literature retrieval, the proposed module does not aim to exhaustively cover the entire literature universe, but instead focuses on collecting evidence that is semantically relevant to the input research topic and potentially useful for law discovery, including quantitative descriptions, experimentally verifiable conclusions, and mechanistic explanations. Such a design improves retrieval efficiency, but it may also introduce retrieval bias. In particular, the topic expansion process and LLM-guided scoring mechanism may favor certain branches that are more frequently discussed or more easily expressed in the existing literature, while under-exploring less common but potentially valuable directions. To mitigate this issue, the system initializes the search from multiple topic keywords, explores diverse branches through the MCTS-based topic expansion mechanism, and retains literature from different semantic paths rather than relying on a single dominant query formulation. Therefore, the retrieval module is intended to improve coverage and diversity of candidate evidence, but it does not eliminate retrieval bias completely. The resulting literature set should be regarded as a structured and topic-oriented evidence pool rather than an unbiased sample of the full scientific literature.

The literature identified in this way is then organized into a structured evidence set, which is represented as a JSON-based literature collection \(\mathcal{L}\), where each entry contains the document title and the corresponding abstract.
The system extracts relevant information such as document titles, abstracts, research subjects, key mechanisms, and verifiable metrics. This structured evidence set forms a comprehensive body of literature directly related to the research topic, providing a reliable foundation for the subsequent hypothesis generation and experimental validation modules. This facilitates the automated research process of literature-hypothesis-experiment.

A related methodological concern is whether the generated hypotheses reflect genuine evidence-guided discovery or merely the recall of patterns already encoded in the language model \cite{illusions,illusions2,qa_dataset}. 
This concern is relevant to AI for science settings, where well-known empirical regularities, such as power-law-like relationships, may already appear in pretraining corpora. 
To reduce such effects, TrafficSci does not rely on the language model to directly output scientific laws from parametric memory. 
Instead, the model organizes, summarizes, and recombines evidence retrieved from the literature collection \(\mathcal{L}\), so that hypotheses are constrained by reported mechanisms, variables, and quantitative observations. 
Although this design cannot fully eliminate prior bias, TrafficSci should be understood as an evidence-guided hypothesis generation framework rather than a strict tabula rasa discovery process. 
The subsequent observational--interventional validation module further subjects each hypothesis to empirical testing, distinguishing evidence-supported hypotheses from unsupported or merely memorized conjectures.

\subsection{Traffic law induction module}

The traffic law induction module aims to generate, screen, refine, and prioritize traffic science hypotheses under explicit theoretical and empirical constraints, transforming free-form language generation into a structured and auditable hypothesis-induction process. Its role is not to provide empirical validation, which is conducted by the observational--interventional validation module, but to prepare candidate hypotheses for validation through evidence anchoring, critic--judge refinement, competitive ranking, and validation-route tagging. The detailed architecture of the module is illustrated in Fig.~\ref{fig:TrafficSci_detailed_design}b.

Given the retrieved literature set \(\mathcal{L}\) and the target research topic \(T\), the hypothesis generation agent first extracts recurring traffic variables, reported mechanisms, and candidate relationships from the retrieved abstracts and metadata. The initial hypothesis generation process is expressed as
\begin{equation}
H_0=\mathrm{HGA}(\mathcal{L},T),
\end{equation}
where \(\mathrm{HGA}(\cdot)\) denotes the hypothesis generation agent and \(H_0\) represents the initial set of candidate hypotheses. Each candidate hypothesis is required to specify the involved variables, the hypothesized relationship, the applicable traffic context, and the type of evidence needed for validation.

To constrain the refinement process, TrafficSci anchors each candidate hypothesis to a subset of retrieved evidence. For a hypothesis \(H\), the system extracts a keyword set \(W_H\), including variables, mechanisms, and traffic phenomena. For each document \(d\in \mathcal{L}\), a corresponding keyword set \(W_d\) is extracted from the title, abstract, and metadata. The relevance between \(H\) and \(d\) is measured by
\begin{equation}
S_{\mathrm{rel}}(H,d)=
\frac{|W_H\cap W_d|}{|W_H\cup W_d|}.
\end{equation}
The top-\(k\) documents with the highest \(S_{\mathrm{rel}}(H,d)\) values are retained as the hypothesis-specific evidence set \(E_H\). Here \(S_{\mathrm{rel}}(H,d)\) is used only for evidence anchoring, not for ranking candidate hypotheses. The resulting evidence set \(E_H\) provides a traceable basis for checking whether the variables, mechanisms, and boundary conditions used in the hypothesis are supported by retrieved literature.

The refinement process is implemented as an evidence-grounded critic--judge loop between two specialized agents: a critic agent and a judge agent. This loop is used as a pre-validation refinement step rather than as empirical validation of a hypothesis.

The critic agent examines each candidate hypothesis \(H_t\) against its hypothesis-specific evidence set \(E_H\). It identifies unsupported variables, inconsistent traffic mechanisms, missing boundary conditions, and insufficient operationalization for empirical testing. To reduce subjective or unconstrained critique, substantive objections are required to refer to specific evidence in \(E_H\), such as retrieved document identifiers, reported variables, or mechanism descriptions. When a hypothesis contradicts basic traffic principles or cannot be translated into measurable operational variables, the critic agent marks the issue as a fatal flaw; otherwise, it provides targeted revision suggestions.

The judge agent then assigns each candidate hypothesis \(H_t\) a four-dimensional quality vector,
\begin{equation}
\mathbf{s}(H_t)=
(s_{\mathrm{val}},s_{\mathrm{nov}},s_{\mathrm{sig}},s_{\mathrm{spe}}),
\end{equation} 
where each component is scored on a discrete scale from 1 to 10.
The four dimensions are defined as follows. \(s_{\mathrm{val}}\) measures consistency with retrieved evidence and basic traffic constraints. \(s_{\mathrm{nov}}\) measures conceptual novelty, namely whether the hypothesis goes beyond a trivial recombination or direct restatement of known variables. \(s_{\mathrm{sig}}\) measures potential relevance for traffic theory, modeling, or management. \(s_{\mathrm{spe}}\) measures whether the variables, boundary conditions, and validation requirements are concrete enough for empirical testing. Based on the critic feedback and judge scores, the hypothesis generation agent iteratively revises the current hypothesis accordingly:
\begin{equation}
H_{t+1}=\mathrm{HGA}(H_t,C_t,\mathbf{s}(H_t)),
\end{equation}
where \(C_t\) denotes the evidence-grounded critique at iteration \(t\). The revised hypothesis is then re-evaluated by the critic--judge loop. This iterative process continues until the hypothesis quality stabilizes, a maximum number of refinement rounds is reached, or the hypothesis satisfies the predefined acceptance criteria for empirical validation. In this process, the critic agent provides qualitative refinement signals, while the judge agent converts the refined hypothesis into four explicit scores.

Candidate hypotheses are selected through competitive elimination. In the implemented ranking procedure, each hypothesis is scored by the arithmetic mean of the four judge scores,
\begin{equation}
S_{\mathrm{comp}}(H)=\frac{1}{4}\left(s_{\mathrm{val}}+s_{\mathrm{nov}}+s_{\mathrm{sig}}+s_{\mathrm{spe}}\right).
\end{equation}
This equal-weight score provides a transparent screening rule that requires a hypothesis to be plausible, non-trivial, relevant and specific before it is passed to empirical validation. We use fixed equal weights to keep the ranking rule inspectable across case studies.

For auditability, the evidence-anchoring scores \(S_{\mathrm{rel}}(H,d)\) are retained with the selected hypothesis to indicate which retrieved documents support the hypothesis-specific evidence set. These relevance scores trace the evidence base of a hypothesis, but they are not used as standalone indicators of scientific novelty and are not included in \(S_{\mathrm{comp}}(H)\).
For each candidate hypothesis, the evidence anchoring step retains the top-\(k\) retrieved documents according to \(S_{\mathrm{rel}}(H,d)\), where \(k\) is fixed across case studies.

Before passing the selected hypothesis to the validation module, TrafficSci assigns a validation-route tag according to the structured content of the hypothesis. If a hypothesis describes a naturally observed statistical pattern, such as a distributional law, scaling relationship, temporal correlation, spatial heterogeneity, or cross-city consistency, it is assigned an observational validation tag. If a hypothesis concerns the effect of a controllable management action, policy variable, control strategy, or counterfactual deployment level, it is assigned an interventional simulation tag. When both descriptive regularity and controllable intervention are involved, it is assigned a combined validation tag. This tag does not itself validate the hypothesis; it only specifies the type of empirical workflow that the subsequent validation module should generate.

The selected hypothesis is passed to the observational--interventional validation module together with its evidence set, critic comments, judge scores, revision history, composite score, variables, boundary conditions, and validation-route tag. This handoff connects the traffic law induction module with the subsequent empirical validation module. The induction module determines whether a hypothesis is evidence-anchored, sufficiently specific, and worth testing; the validation module then executes the corresponding observational or simulation-based experiment to evaluate empirical support.

\subsection{Observational--interventional validation module}
The observational--interventional validation module consists of a procedure agent and an experiment agent. This module validates the structured hypotheses generated by the traffic law induction module by converting them into executable experimental workflows, forming an automated loop from hypothesis construction to experimental validation and hypothesis updating. It autonomously selects experimental methods, generates experimental steps, calls data sources and external tools, and performs validation tasks, thereby reducing human involvement and improving the efficiency of traffic science research \cite{llmtool,swe}.

This module follows two complementary validation paradigms: observational validation and interventional validation. Observational validation tests hypotheses using naturally collected real-world traffic data without actively changing the traffic system. It is suitable for examining empirical regularities such as distributional patterns, temporal correlations, spatial heterogeneity, and cross-city consistency. Interventional validation tests hypotheses by actively manipulating key variables in traffic simulation environments, making it suitable for evaluating control-related or policy-related effects that are difficult, costly, or unsafe to test directly in the real world. To formalize this closed-loop process, the observational--interventional validation module is represented as an operator:
\begin{equation}
\label{eq:automated_loop}
(H_{t+1},\, r_t) \;=\; \mathcal{A}(H_t),
\end{equation}
where $H_t$ denotes the structured hypothesis at iteration $t$, $r_t$ is the structured validation result, including dataset sources, experimental methods, key metrics, visualizations, and validation conclusions, and $H_{t+1}$ is the updated hypothesis obtained by feeding the validation result $r_t$ back to the traffic law induction module for refinement, thereby closing the discovery loop.

As illustrated in Fig.~\ref{fig:TrafficSci_detailed_design}c, the procedure agent first receives the structured hypotheses, including variables, relationships, and applicable conditions. It analyzes the validation requirements and determines whether each hypothesis should be tested through observational analysis, interventional simulation, or a combination of both. For example, a hypothesis about a heavy-tailed congestion cost distribution can be mapped to an observational workflow involving data selection, distribution fitting, and scaling exponent estimation, whereas a hypothesis about traffic control penetration can be mapped to an interventional workflow involving simulation scenario construction, controlled variable manipulation, and intervention analysis.

The experiment agent then converts the procedure plan into executable experimental steps, including data selection, preprocessing, method selection, metric design, tool invocation, and result organization. For observational validation, relevant datasets are retrieved from internal repositories or external data sources, followed by targeted preprocessing such as missing-value imputation, denoising, temporal aggregation, and variable extraction. For interventional validation, simulation scenarios are constructed in traffic simulators such as SUMO, where traffic demand, control strategies, penetration rates, or other key variables can be actively manipulated. The system then records changes in traffic indicators, such as travel time, congestion level, traffic efficiency, and emissions, to evaluate whether the hypothesis is supported.

The system supports various validation tools, including statistical analysis, time-series analysis, regression analysis, distribution fitting, and simulation-based evaluation \cite{toolformer}. Since traffic hypotheses differ substantially in data form, spatio-temporal scale, and evaluation criteria, a fixed experimental template is insufficient. To address this challenge, the procedure agent dynamically decomposes each hypothesis into validation objectives, required variables, applicable datasets, and candidate methods, while the experiment agent executes the corresponding workflow with suitable tools and metrics.

To integrate heterogeneous validation tools and traffic-specific environments, the model context protocol (MCP) is adopted as a standardized interface for tool integration, enabling access to external environments and services through a unified protocol. Built upon MCP, a modular skills framework is developed to encapsulate reusable traffic validation workflows, such as distribution fitting, scaling-law estimation, temporal correlation analysis, regression testing, scenario generation, and SUMO-based intervention evaluation. These skills can be composed into end-to-end executable pipelines for different types of traffic hypotheses.

Finally, the experiment agent generates a structured validation report, including dataset sources, preprocessing operations, experimental methods, parameter settings, quantitative metrics, visualizations, and validation conclusions. These results are fed back to the traffic law induction module, allowing each hypothesis to be refined, rejected, or generalized based on empirical evidence, thereby closing the discovery loop. This design enables TrafficSci to combine real-world observational evidence with simulation-based interventional evidence, providing a traffic-specific experimental foundation for automated discovery and verification of traffic laws.

\section*{Acknowledgements}

This work was supported in part by the National Natural Science Foundation of China under Grants 62271485 and 62303462, and in part by the Science and Technology Development Fund, Macao Special Administrative Region under Grants 0093/2023/R1A2, 0145/2023/R1A3 and 0157/2024/R1A2. 
During manuscript preparation, the authors used ChatGPT and Claude for language polishing. The authors reviewed and revised all outputs and take full responsibility for the final manuscript.

\section*{Data availability}
 The mobility data used in the universal visitation law analysis across four cities are publicly available at https://github.com/leiii/VisitationLaw. The data used for the congestion-cost analysis based on jam-prints are publicly available at https://github.com/GuanwenZeng/Jam-prints. The data and simulation settings used for the adaptive-signal-control analysis are available within the original study cited in the Article and its Zenodo repository at https://doi.org/10.5281/zenodo.14591154. The trajectory data used for the temporal-memory analysis are publicly available from Argoverse 2 at https://www.argoverse.org/av2.html and nuScenes at https://www.nuscenes.org/nuscenes under their respective terms of use. 
\section*{Code availability}
 The source code implementing TrafficSci and reproducing the analyses in this study will be made publicly available upon acceptance of the manuscript. 

\section*{Author contributions}
Y.Lv, X.D. and F.-Y.W. designed the research; X.D., Y.Liu, X.G. and Q.M. performed the research; Y.Liu, J.S. and Y.C. analyzed the data; X.D., Y.Liu, X.G., Q.M., J.S., Y.W., C.G., Y.T., Y.C., C.X., Y.Lv and F.-Y.W. wrote the paper.

\section*{Competing interests}

The authors declare no competing interests.
\bibliography{references}

\end{document}